\documentclass[unnumsec,webpdf,contemporary,large]{oup-authoring-template}%
\usepackage[table]{xcolor}
\usepackage{booktabs}
\usepackage{amsfonts}
\usepackage{amsmath}
\usepackage{multirow}
\usepackage{tabularx}
\usepackage{bm}

\graphicspath{{Fig/}}

\theoremstyle{thmstyleone}%
\theoremstyle{thmstyletwo}%
\theoremstyle{thmstylethree}%

\usepackage[T1]{fontenc}
\usepackage[utf8]{inputenc}
\begin{document}

\journaltitle{Journal Title Here}
\DOI{DOI HERE}
\copyrightyear{2026}
\pubyear{2026}
\access{Advance Access Publication Date: Day Month Year}
\appnotes{Paper}

\firstpage{1}


\title[Short Article Title]{Local–Global Multimodal Contrastive Learning For Molecular Property Prediction}

\author[1]{Xiayu Liu \ORCID{0009-0001-5333-0754}}
\author[2]{Zhengyi Lu \ORCID{0009-0003-4723-7187}}
\author[3]{Yunhong Liao \ORCID{0009-0004-9765-0065}}
\author[4]{Chao Fan \ORCID{0000-0002-2435-6656}}
\author[1,$\ast$]{Hou-biao Li\ORCID{0000-0002-7268-307X}}

\authormark{Author Name et al.}

\address[1]{\orgdiv{School of Mathematical Sciences}, \orgname{University of Electronic Science and Technology of China}, \orgaddress{\street{No.2006, Xiyuan Avenue, West Hi-tech Zone}, \postcode{611731}, \state{Chengdu}, \country{China}}}
\address[2]{\orgdiv{Department of Computer Science and Engineer}, \orgname{Oakland University}, \orgaddress{\street{201 Meadow Brook RD, Rochester Hills}, \postcode{48309}, \state{MI}, \country{USA}}}
\address[3]{\orgdiv{Department of Electrical and Computer Engineering}, \orgname{Oakland University}, \orgaddress{\street{201 Meadow Brook Rd, Rochester Hills}, \postcode{48309}, \state{MI}, \country{USA}}}
\address[4]{\orgdiv{College of Management Science}, \orgname{Chengdu University of Technology}, \orgaddress{\street{No.1, East Third Road, Erxianqiao, Chenghua District}, \postcode{610051}, \state{Chengdu}, \country{China}}}

\corresp[$\ast$]{Corresponding author: Hou-biao Li, \href{email:lihoubiao0189@163.com}{lihoubiao0189@163.com}}

\received{Date}{0}{Year}
\revised{Date}{0}{Year}
\accepted{Date}{0}{Year}

\abstract{Accurate molecular property prediction requires integrating complementary information from molecular structure and chemical semantics. In this work, we propose LGM-CL, a local-global multimodal contrastive learning framework that jointly models molecular graphs and textual representations derived from SMILES and chemistry-aware augmented texts. Local functional group information and global molecular topology are captured using AttentiveFP and Graph Transformer encoders, respectively, and aligned through self-supervised contrastive learning. In addition, chemically enriched textual descriptions are contrasted with original SMILES to incorporate physicochemical semantics in a task-agnostic manner. During fine-tuning, molecular fingerprints are further integrated via Dual Cross-attention multimodal fusion. Extensive experiments on MoleculeNet benchmarks demonstrate that LGM-CL achieves consistent and competitive performance across both classification and regression tasks, validating the effectiveness of unified local-global and multimodal representation learning.}

\maketitle
\section{Introduction}
Molecular property prediction plays a fundamental role in modern drug discovery and chemical research, as it enables the efficient estimation of physicochemical, biological, and toxicological properties of candidate molecules prior to costly experimental validation. Accurate prediction of molecular properties such as bioactivity, solubility, permeability, toxicity, and binding affinity can significantly accelerate the early stages of drug development, reduce experimental costs, and improve the success rate of lead compound identification.\par

Early computational approaches to molecular property prediction primarily relied on handcrafted molecular descriptors and fingerprints, including physicochemical descriptors and substructure-based features \cite{jorissen2005virtual,mahe2006pharmacophore,svetnik2003random}. These representations were typically combined with conventional machine learning algorithms such as Support Vector Machines (SVM) \cite{cortes1995support}, and Random Forests (RF) \cite{breiman2001random}. While such methods achieved encouraging performance on various tasks, their predictive capability heavily depends on manually crafted descriptors, which often fail to fully capture complex chemical interactions. Consequently, these approaches may struggle to generalize across chemically diverse prediction scenarios\par

With the rapid advancement of deep learning, a wide range of neural architectures have been proposed for molecular property prediction \cite{duvenaud2015convolutional}. In contrast to traditional approaches, deep models can automatically learn task-specific molecular representations directly from raw molecular structures. In particular, graph neural networks (GNNs) have gained widespread adoption due to their natural compatibility with molecular graph structures, where atoms and bonds are represented as node and edges, respectively. Message passing paradigms \cite{gilmer2017neural,kipf2016semi,hamilton2017inductive,velivckovic2017graph,xu2018powerful} iteratively aggregate local chemical information to generate molecular embeddings. Meanwhile, transformer architectures have been introduced to molecular representation learning to better address long-range and global dependencies through self-attention mechanisms \cite{chen2023graph,luo2022one,ying2021transformers,vaswani2017attention}. Despite their effectiveness, existing approaches still face challenges in jointly modeling molecular information across different structural scales.\par

Molecular properties are intrinsically governed by chemical patterns at multiple structural scales. Local chemical environments, such as functional groups and atom-level neighborhoods, determine reactivity and local chemical behavior \cite{morgan1965generation}, whereas global molecular context, including long-range dependencies and overall topology, influences macroscopic properties such as stability, solubility, and bioactivity \cite{gasteiger2020directional}. However, existing molecular representation learning methods often emphasize either local or global structure. Localized message passing approaches excel at capturing neighborhood level interactions but may inadequately encode global dependencies, while models designed for long-range modeling may overlook subtle local chemical patterns. This imbalance motivates the explicit integration of local and global molecular information as complementary components for molecular representation learning.\par

Beyond graph-based representations, molecular properties are also closely related to chemical patterns encoded in molecular strings. The Simplified Molecular Input Line Entry System (SMILES) \cite{weininger1988smiles} provides a linearized description of molecular structure, preserving sequential patterns and implicit chemical rules. Inspired by advances in natural language processing, SMILES strings have been increasingly utilized as an alternative representation for molecular property prediction using transformer-based language models \cite{chithrananda2020chemberta}. However, SMILES representations primarily capture structural syntax and may offer a limited view of higher level physicochemical attributes. Recently, large language models (LLMs) have been explored to generate chemically relevant textual descriptions as  auxiliary molecular representations \cite{zhang2024molecular,guo2024moltailor,zheng2023large}. While promising, the application of LLM-generated content introduces the risk of hallucination, where chemically incorrect or unsupported information may be produced \cite{xu2024hallucination}. Mitigating hallucination  therefore requires constraining the generation process through chemistry-aware prompting strategies, enabling LLM-generated text to serve as a reliable and complementary view.\par

Taken together, these observations suggest that molecular property prediction is inherently a multiview problem \cite{zhang2024pre,liu2024git}, where different representation modalities capture complementary aspects of molecular information. Graph-based representations emphasize molecular topology and multiscale structural interactions, while sequence- and text-based representations provide additional semantic perspectives derived from linearized structure and chemically informed knowledge. Relying on any single modality is therefore insufficient to fully characterize complex structure-property relationships\par

Learning unified representation from heterogeneous molecular modalities remains challenging problem due to differences in feature distributions and inductive biases. In this context, contrastive learning provides a principled framework for multimodal representation learning by aligning multiple views of the same molecule in a shared embedding without explicit supervision \cite{sun2021mocl,chen2020simple,radford2021learning}. By encouraging consistency across complementary modalities, contrastive objectives enable the learning of robust and transferable molecular representations while mitigating the limitations of individual views.\par
To address the above challenges, we propose LGM-CL, a unified Local–Global Multimodal Contrastive Learning framework for molecular property prediction. Starting from SMILES strings, LGM-CL constructs multiple complementary molecular views that capture local structural patterns, global topological context, and physicochemical knowledge at different levels. Structural information is modeled from molecular graphs to characterize both local functional group environments and global molecular organization, while textual information is derived from SMILES strings and chemically informed descriptions to encode physicochemical characteristics beyond pure structural syntax.\par
LGM-CL adopts a self-supervised multiview contrastive learning strategy to effectively align heterogeneous molecular representations without relying on task-specific supervision. By contrasting different structural and textual views of the same molecule, the framework learns consistent and transferable molecular embeddings that integrate local and global structure cues with chemically meaningful semantic information. For downstream molecular property prediction, the learned representations are further refined by incorporating molecular fingerprints as an additional complementary modality, enabling unified fusion of structural, semantic, and fingerprint-based information.\par
The main contributions of this work are summarized as follows:
\begin{itemize}
\item We propose a chemistry-aware prompting template for SMILES augmentation, which reduces hallucination in LLM-generated text and enables self-supervised contrastive learning between original SMILES and chemically enriched textual descriptions.
\item We introduce a local–global contrastive graph representation learning scheme, where molecular graphs are encoded by separate local and global encoders to jointly capture functional group information and global topological context.
\item We develop a unified multi-modal framework (LGM-CL) that integrates textual, graph-based, and molecular fingerprint representations to learn transferable molecular embeddings for property prediction.
\item Extensive experiments on multiple benchmark datasets demonstrate the effectiveness of LGM-CL, with ablation studies and visualization analyses validating the contribution of each component.

\end{itemize}

\section{Materials and methods}\label{sec2}
\subsection{Overview of Model}\label{subsec1}
Our framework aims to learn a chemically meaningful and transferable molecular representation by unifying local function group patterns, global topological context and physicochemical semantics derived from SMILES and LLM-enhanced textual descriptions, through self-supervised multi-view contrastive learning and cross-modal attention-based fusion. Starting from SMILES strings, we construct multiple complementary molecular views that capture structural and semantic information at different levels. Specifically, we build a graph-based view that explicitly models both local neighborhood structures and global topological dependencies by employing distinct graph encoders, and a text-based view constructed from original SMILES strings together with chemistry-aware textual augmentations that emphasize physicochemical semantics.\par
We first apply a set of self-supervised multi-view contrastive learning objectives over these representations, including view-specific objectives for the graph and text modalities, as well as cross-view objectives that encourage consistency between complementary representations of the same molecule. For the graph modality, different structural views derived from the same molecular graph are jointly contrasted to emphasize both local function group information and global molecular context. For the text modality, chemically enriched descriptions generated from SMILES are contrasted with their original string-based representations, enabling the textual encoder to capture domain-specific physicochemical semantics beyond pure syntactic structure. Through this multi-view contrastive pretraining, the model learns transferable molecular representations that integrate local and global structural information with chemically meaningful semantic knowledge.\par
\begin{figure*}[t]
    \centering
    \includegraphics[width=\textwidth]{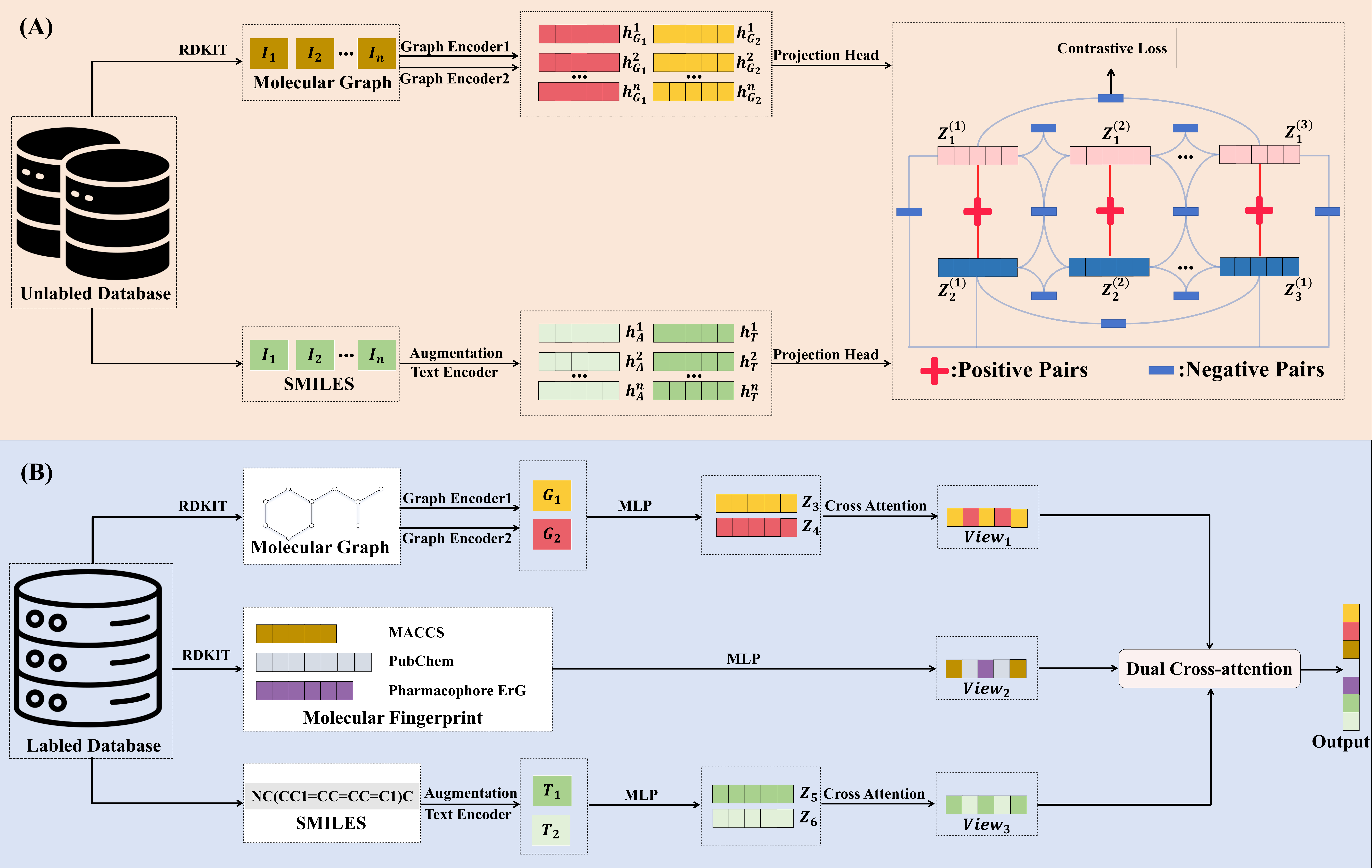}
    \caption{Overview of Model. (A) Pipeline of the model in the pre-training stage. (B) Pipeline of the model when applied to downstream tasks.}
    \label{fig:Overview}
\end{figure*}

Following the multi-view contrastive pretraining stage, the model is further adapted to downstream tasks in a fine-tuning stage by incorporating molecular fingerprint information as an additional complementary modality. In this stage, pretrained graph- and text-based representations are first consolidated within each modality to form unified graph and text embeddings, and are then jointly aligned with fingerprint-based features through a Dual Cross-attention module to obtain a single molecular representation for prediction. This hierarchical integration strategy enables effective fusion of structural, semantic, and fingerprint-derived information while preserving their complementary roles. An overview of the model is provided in Figure ~\ref{fig:Overview}.
\subsection{Graph Modality Contrastive Learning}
\begin{figure*}[t]
    \centering
    \includegraphics[width=\textwidth]{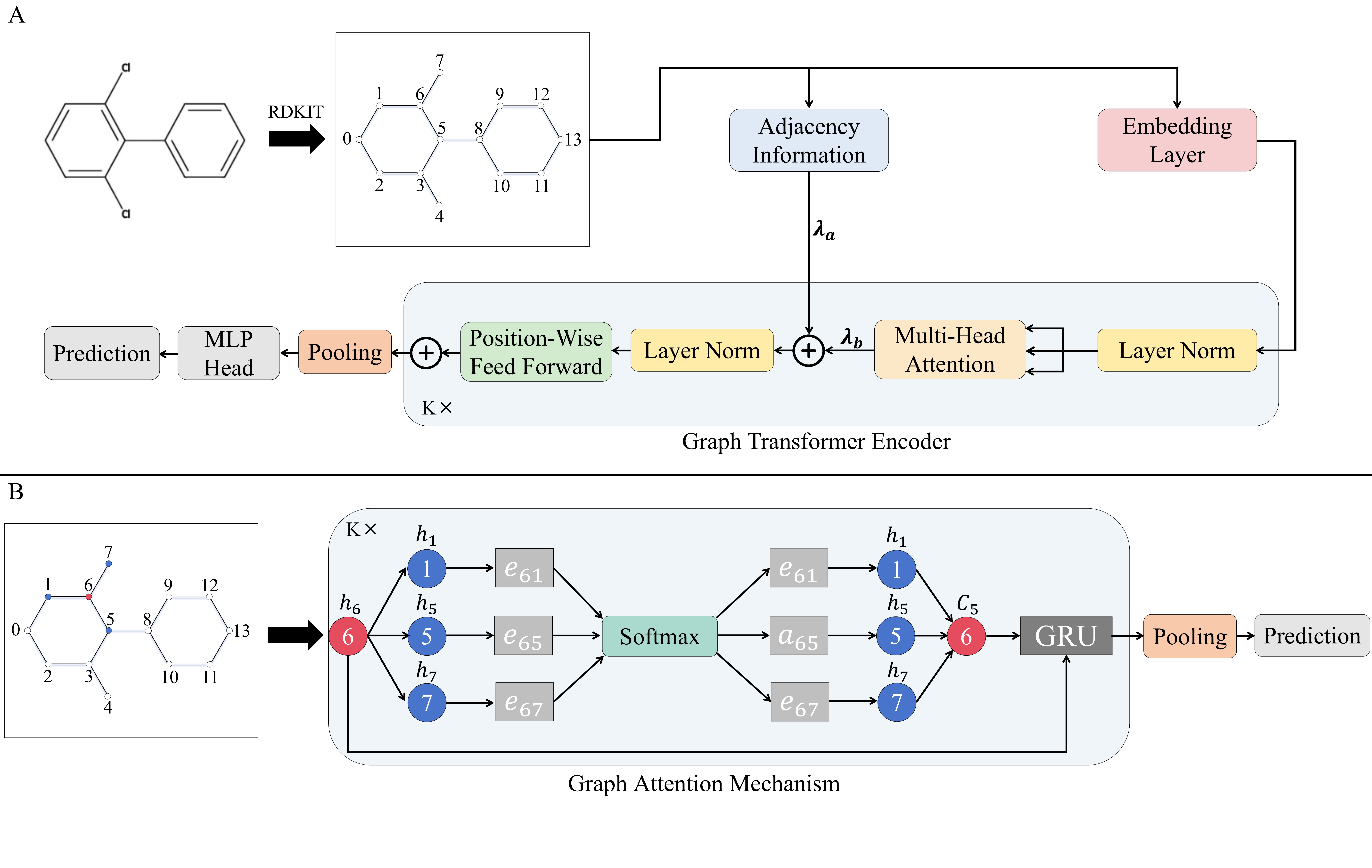}
    \caption{Overview of global and local graph encoders. (A) The Graph Transformer encoder, which model global structural dependencies of molecular graphs through multi-head self-attention with adjacency-aware bias. (B) The Attentive FP encoder, which captures local chemical environments via attention-guided message passing followed by a GRU-based update and pooling operation.}
    \label{fig:GT and AT}
\end{figure*}

Inspired by the dual-view contrastive learning paradigm proposed in DGCL \cite{jiang2024dgcl}, we design a graph modality contrastive learning strategy that explicitly captures molecular information at different structural scales. To this end, we employ two structurally complementary graph encoders to generate distinct yet semantically consistent representations of the same molecular graph. The two encoders are designed to focus on complementary structural scales of the molecular graph, corresponding to a global view and a local view, respectively, as described below. An overview of the two encoders is illustrated in Figure ~\ref{fig:GT and AT}(A) and Figure ~\ref{fig:GT and AT}(B), respectively.
\subsubsection{Global Graph Encoder via Graph Transformer}
Accurate molecular property prediction often requires modeling global structural dependencies, as small local modifications may lead to substantial changes in molecular activity, a phenomenon commonly referred to as activity cliffs. Capturing such long-range interactions and holistic molecular context is therefore critical for robust prediction. Motivated by this observation and following prior work \cite{maziarka2020molecule, maziarka2024relative, yun2019graph, liu2025multilevel}, we adopt a graph transformer as a global graph encoder to account for long-range dependencies and global molecular structure.
Given an arbitrary SMILES sequence $S$, we first convert it into a molecular graph $G=(V,E)$ using RDKIT, where $V$ and $E$ denote the sets of atoms and bonds, respectively. Each atom $v_{i}\in V$ and bond $e_{k}\in E$ is initialized with a set of chemical descriptors according to a predefined encoding scheme, as summarized in Table~\ref{tab:features}. This process yields the initial atom features $\{a_i\}_{i=1}^{n}$ and bond features $\{b_{k}\}_{k=1}^{m}$, where $n$ and $m$ denote the numbers of atoms and bonds in the molecule. The atom and bond features are then jointly embedded through an embedding layer, producing a node feature matrix $M\in R^{n\times d}$, where each row corresponds to a contextualized atom representation. Based on $M$, linear projections are applied to obtain the query, key, and value matrices $Q$, $K$, and $V$, which are subsequently used to compute attention scores according to the graph transformer formulation. By integrating node feature interactions with adjacency-aware attention, the resulting representations enable effective propagation of information across the entire molecular graph, facilitating the modeling of long-range dependencies and global molecular structure. Specifically, the attention output for each node is computed as 
\begin{equation}
\label{attention_scores}
    A_{score}=(\lambda_{a}\cdot softmax(\frac{QK^{T}}{\sqrt{d_{k}}}))V
\end{equation}
Specifically, Eq.\ref{attention_scores} enables each node to aggregate information from all other nodes in the molecular graph through attention-weighted interactions. The scaled dot-product attention computes pairwise relevance scores between atoms, allowing the model to capture long-range structural dependencies beyond local neighborhoods. By integrating information across the entire graph, the Graph Transformer encoders global molecular context, which is essential for modeling activity cliffs where small local changes can induce substantial global effects.
\begin{table*}[t]
\renewcommand{\arraystretch}{1.2}
\caption{Initial Atomic and Bond Features.}
\label{tab:features}
\tabcolsep=0pt
\begin{tabular*}{\textwidth}{@{\extracolsep{\fill}}l c p{0.7\textwidth}@{\extracolsep{\fill}}}
\toprule
\rowcolor{gray!10}
\textbf{Atom Features} & \textbf{Size} & \textbf{Description} \\
\midrule
Atom Symbol       & 16 & [C, N, O, F, Si, Cl, As, Se, Br, Te, I, At, others] (one-hot) \\
Degree            & 6  & Number of atoms connected to the atom (one-hot) \\
Formal Charge     & 1  & Electrical charge (integer) \\
Radical Electrons & 1  & Number of radical electrons (integer) \\
Hybridization     & 6  & [sp, sp\textsuperscript{2}, sp\textsuperscript{3}, sp\textsuperscript{3}d, sp\textsuperscript{3}d\textsuperscript{2}, other] (one-hot) \\
Aromaticity       & 1  & Whether the atom is part of an aromatic system (one-hot) \\
Hydrogens         & 5  & Number of hydrogens connected to the atom (one-hot) \\
Chirality Type    & 4  & The chirality type of the atom (one-hot) \\
Ring              & 1  & Whether the atom is in a ring (one-hot) \\
Ring Type         & 4  & [3, 4, 5, 6] (one-hot) \\
Atomic Mass       & 1  & Mass of the atom (integer) \\
Implicit Valence  & 7  & [0, 1, 2, 3, 4, 5, 6, 7] (one-hot) \\
Hydrogen Acceptor & 1  & Whether the atom acts as a hydrogen acceptor (one-hot) \\
Hydrogen Donor    & 1  & Whether the atom acts as a hydrogen donor (one-hot) \\
Acidic            & 1  & Whether the atom is part of an acidic group (one-hot) \\
Basic             & 1  & Whether the atom is part of a basic group (one-hot) \\
\midrule
\rowcolor{gray!10}
\textbf{Bond Features} & \textbf{Size} & \textbf{Description} \\
\midrule
Bond Type   & 5 & [exists, single, double, triple, aromatic] (one-hot) \\
Conjugation & 1 & Whether the bond is conjugated (one-hot) \\
Wedge Bond & 2 & [endupright, enddownright](one-hot)\\
Ring        & 1 & Whether the bond is in a ring (one-hot) \\
Stereo      & 6 & [0, 1, 2, 3, 4, 5] (one-hot) \\
\bottomrule
\end{tabular*}
\end{table*}

\subsubsection{Local Graph Encoder via Attentive FP}
while global structural context plays an important role in molecular property prediction, many molecular properties are strongly influenced by local chemical environments, such as functional groups and other chemically meaningful molecular motifs.
These local structures often determine key physicochemical characteristics and reactivity patterns, making their accurate modeling essential for reliable prediction. To capture such localized information, we employ Attentive FP \cite{xiong2019pushing} as the local graph encoder.\par
Attentive FP updates atom representations through an attention-guided message passing mechanism \cite{gilmer2017neural} that focuses on short-range chemical interactions. In our framework, we intentionally adopt a shallow Attentive FP architecture with a small number of message passing layers, encouraging the encoder to emphasize local atomic neighborhoods and function-group-level patterns, rather than propagating information over long graph distances that are already modeled by the global graph encoder.\par
Following the initial feature embedding described in the previous section, each atom is associated with a contextualized node representation, forming a node feature matrix
\begin{equation}
    M=[h_{1}^{(0)},h_{2}^{(0)},...,h_{n}^{(0)}]^{T}\in R^{n\times d}
\end{equation}
where $h_{v}^{(0)}$ denotes the initial representation of atom $v$. At message passing layer $l$, an attention score is first computed for each pair of neighboring atoms $(v,u)$. Specifically, the unnormalized attention coefficient is defined as
\begin{equation}
    e_{vu}^{(l-1)}=LeakyReLU(W_{vu}^{(l-1)} [h_{v}^{(l-1)},h_{u}^{(l-1)}])
\end{equation}
which measures the importance of neighbor atom $u$ to atom $v$ based on their representations at the previous layer. The attention coefficients are then normalized over the neighborhood $N(v)$ using a softmax function:
\begin{equation}
    a_{vu}^{(l-1)}=softmax(e_{vu}^{(l-1)})=\frac{exp(e_{vu}^{(l-1)})}{\sum_{i\in N(v)}exp(e_{vi}^{(l-1)})}
\end{equation}
ensuring that the contributions from all neighboring atoms sum to one. Using these attention weights, a context vector for atom $v$ is computed as a weighted aggregation of its neighbors:
\begin{equation}
    c_{v}^{(l-1)}=ELU(\sum_{u\in N(v)} a_{vu}^{(l-1)}\cdot W_{v}^{(l-1)}h_{u}^{(l-1)})
\end{equation}
where the attention mechanism allows the model to selectively emphasize chemically relevant neighboring atoms and local substructures. Finally, the atom representation is updated via a gated recurrent unit (GRU) \cite{chung2014empirical}:
\begin{equation}
    h_{v}^{(l)}=GRU(c_{v}^{(l-1)},h_{v}^{(l-1)})
\end{equation}
which integrates the newly aggregated local context while preserving previously learned atomic information. Through a limited number of such message passing layers, Attentive FP produces refined atom representations that encode rich local chemical environments, providing an effective local graph encoder complementary to the global structural component.

\subsubsection{Contrastive Alignment of Global and Local Graph Representations}
To jointly align global and local structural information, we introduce a contrastive learning objective between the representations learned by the global and local graph encoders. Given a molecular graph, the two encoders process the same input structure but are designed to capture complementary information at different structural scales. Specifically, AttentiveFP is employed as a local graph encoder with a shallow message passing architecture, which restricts information propagation to a limited neighborhood range. This design encourages the model to focus on local chemical environments, such as functional groups and short-range bonding patterns that are critical for determining local reactivity and physicochemical properties. In contrast, the Graph Transformer serves as a global graph encoder by leveraging a generic self-attention mechanism, enabling each atom to attend to all other atoms in the molecular graph. This formulation allows the model to capture long-range dependencies and holistic molecular context, which are essential for modeling global structural effects and activity cliffs. For each molecular graph, atom-level representations obtained from the two encoders are aggregated via a sum-based readout to form molecule-level and local embeddings.\par
Although both representations are derived from the same molecular graph, they capture complementary structural information at different levels of granularity. To encourage consistency between two views, we perform contrastive learning on the corresponding representation pairs. Specifically, we adopt the Normalized Temperature-scaled Cross entropy(NT-Xent) loss as the contrastive objective. Within a mini-batch of $N$ molecules, the global and local representations of the same molecule form a positive pair, while representations from different molecules serve as negative samples. The contrastive objective aligns local and global structural views, resulting in consistent molecular representations. The NT-Xent loss is defined as:
\begin{equation}
    \begin{aligned}
    \ell&(z_{i},z_{j})=exp(sim(z_{i},z_{j})/\tau)\\
    \mathcal{L}&_{CL}=-\frac{1}{N}\sum_{i=1}^{N}log\frac{\ell(z_{global}^{(i)},z_{local}^{(i)})}{\sum_{k=1}^{N} 1_{[k\ne i]}\ell(z_{global}^{(i)},z_{global}^{(k)})+\ell(z_{global}^{(i)},z_{local}^{(k)})}
    \end{aligned}
    \label{contrastive}
\end{equation}
where $z_{global}$ and $z_{local}$ denote the molecule-level representations obtained from the global and local graph encoders, respectively; $1_{[k\ne i]}$ is an indicator function excluding the self-pair. $sim(\cdot,\cdot)$ denotes the cosine similarity function and $\tau$ is a temperature hyperparameter.\par 
By minimizing this contrastive objective, the global and local representations corresponding to the same molecule are pulled closer in the embedding space, while representations associated with different molecules are pushed apart. This loss enforces cross-scale alignment between global and local graph representations, enabling the model to capture consistent structural semantics across different levels of granularity. As a result, the contrastive objective serves as an effective self-supervised pretraining signal that facilitates robust and informative molecular representation learning.
\subsection{Text Modality Contrastive Learning}
\subsubsection{LLM-Based Semantic Augmentation with Prompt Templates}
In order to enrich the semantic capacity of the textual modality, we augment the raw SMILES strings with chemically meaningful natural language descriptions generated by Mistral-7B-Instruct-v0.3 \cite{Albert2023mistral}. Although SMILES provides an effective symbolic encoding of molecular structure, it lacks explicit information about key chemical characteristics, such as function groups, reactivity patterns, stereochemical features, and broader physicochemical profiles. These semantics attributes, which are crucial for understanding molecular behavior, are not directly accessible from the SMILES content. In response to this limitation, we prompt Mistral-7B-Instruct-v0.3 with a carefully designed template to elicit descriptive textual information that reflects the underlying chemical semantics of the molecule. To guide the language model towards producing chemically reliable and semantically structured descriptions, drawing inspiration from prior studies on prompt engineering for scientifc and chemical language modeling \cite{jin2025effective,li2025gicl},we design a prompt template that incorporates several elements of prompt engineering. The template begins with a role-prompting instruction, explicitly conditioning the large language model to assume the role of an expert medicinal chemist and cheminformatics scientist. This role specification encourages the model to adopt domain-appropriate reasoning patterns, terminology, and representational biases when interpreting a molecule. In addition, the template incorporates strict content constraints motivated by the known tendency of large language models to generate incorrect or fabricated numerical values. To prevent such numerical hallucinations, the model is explicitly instructed not to invent quantitative information. Furthermore, as a measure to avoid introducing task-specific biases that could interfere with downstream property prediction, the template disallows any reference to particular prediction tasks or biological endpoints. Finally, to ensure compatibility with an unsupervised contrastive learning setting, the model is restricted to producing descriptions grounded solely in the SMILES string and the accompanying RDKIT-derived descriptors, without relying on any external or task-dependent information. These constraints serve as safeguards to ensure that the generated text remains factual and chemically interpretable. Because large language models do not always adhere perfectly to a fixed output structure, we further apply a post-processing step that normalizes the generated content into a unified format. This preprocessing ensure that every molecule is represented using a consistent template, which is essential for stable contrastive learning in the subsequent stage. Workflow for enriching SMILES strings with chemically meaningful natural language descriptors and a simplified illustration of the prompt template is shown in Figure ~\ref{fig:Template}, and the complete version is provided in the supplementary material.

\subsubsection{Contrastive Alignment of SMILES and Textual Representation}
To effectively exploit the complementary information present in the two textual views, we perform contrastive learning between the raw SMILES strings and the augmented descriptions generated by the prompt-guided language model. The SMILES representation primarily conveys the molecular topological structure information, including atom connectivity, local structural patterns, and implicit geometric constraints. In contrast, the augmented text provides higher-level chemical semantics that are not explicitly accessible from the  SMILES alone, including the core scaffold, salient structural features, physicochemical tendencies, and general medicinal chemistry attributes elicited through the designed prompt template. By contrasting these two perspectives of the same molecule, the model is encouraged to associate structural cues with their corresponding semantic interpretations, thereby learning a richer and more chemically informative representation. Such alignment facilitates downstream property prediction, as the learned embedding integrates both precise structural information and abstract chemical knowledge that jointly contribute to molecular behavior. \par
\begin{figure*}[t]
    \centering
    \includegraphics[width=\textwidth]{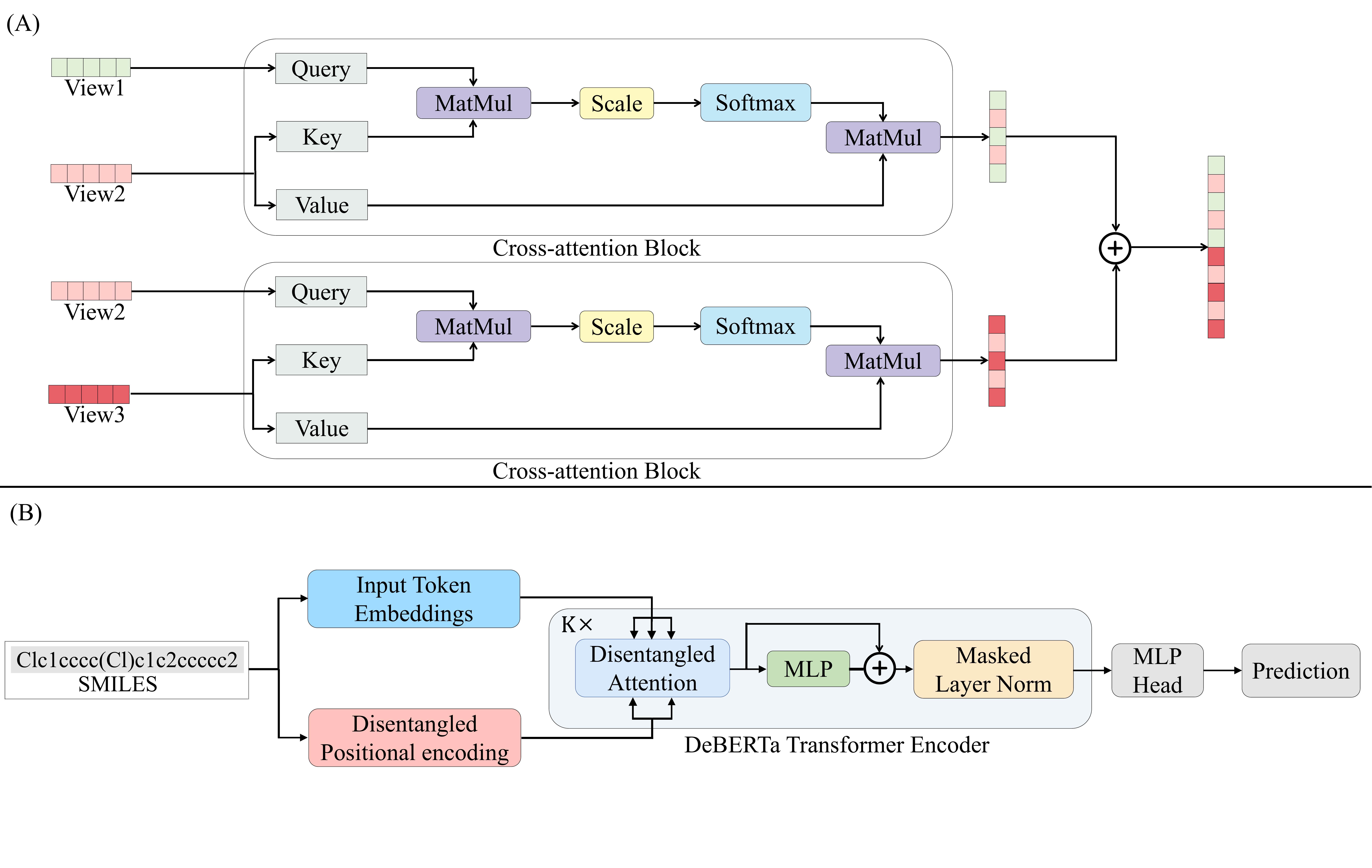}
    \caption{Schematic illustration of the dual cross-attention component and text encoder in our framework. (A) The details of the dual cross-attention component. (B) DeBERTa-based encoder for SMILE and LLM-augmented textual descriptions.}
    \label{fig:DeBERTa and Dual}
\end{figure*}

Implementing this contrastive alignment requires an encoder capable of handing both symbolic SMILES sequences and natural language descriptions with a unified architecture. For this purpose, we utilize DeBERTa \cite{he2020deberta}, whose disentangled attention design enables it to effectively capture view-specific information from each of the two views.. A schematic illustration of the DeBERTa architecture is provided in Figure ~\ref{fig:DeBERTa and Dual}(B).
DeBERTa is a transformer-based language model that improves contextual representation learning through a disentangled attention mechanism, in which content and position information are encoded separately. Given a textual input $X_{i}=\{x_{1},x_{2},...,x_{n}\}$, DeBERTa first maps each token into a continuous representation through an embedding layer, resulting in a content embedding matrix $H_{i}\in R^{n\times d}$, where $n$ denotes the sequence length and d is the hidden dimension. In parallel, relative positional relationships between token pairs are encoded by a relative position embedding module, producing position representations $P_{i}\in R^{2k\times d}$, where k denotes the maximum relative distance considered in the model. Based on the content embeddings $H_{i}$ and the shared relative position representations $P_{i}$, DeBERTa computes attention score using a disentangled formulation that explicitly models content-to-content, content-to-position, and position-to-content interactions, as defined below:
\begin{equation}
    \begin{aligned}
Q&_{c}=H_{i}W_{q,c}, \quad K_{c}=H_{i}W_{k,c}, \quad V_{c}=H_{i}W_{v,c} \\
Q&_{r}=P_{i}W_{q,r}, \quad K_{r}=P_{i}W_{k,r}\\
\tilde{A}&_{i,j} =
\underbrace{Q_{i}^{c} K_{j}^{c\top}}_{\text{(a) content-to-content}}
+
\underbrace{Q_{i}^{c} K_{\delta(i,j)}^{r\top}}_{\text{(b) content-to-position}}
+
\underbrace{K_{j}^{c}Q_{\delta(j,i)}^{r\top}}_{\text{(c) position-to-content}}\\
H&_{i}^{o}=softmax(\frac{\tilde{A}}{\sqrt{3d}})V_{c}
\end{aligned}
\end{equation}
where $W_{q,c},W_{k,c},W_{v,c},W_{q,r},W_{k,r}\in R^{d\times d}$ are learnable projection matrices for content and positional components, respectively. The function $\delta_{(i,j)}$ maps the relative distance between token i and token j to an index in the relative position embedding space. The attention score $\tilde{A}_{i,j}$ is computed as the sum of content-to-content, content-to-position, and position-to-content interactions. The resulting representation $H_{o}$ corresponds to the output features vectors of this attention layer and serves as the input to the subsequent layers or downstream output modules.\par
After encoding the SMILES string and their corresponding LLM-augmented textual descriptions with DeBERTa, we obtain two complementary textual representations for each molecule. To align these two views, we apply contrastive learning that brings representations from the same molecule closer while separating those from different molecules. Following the same NT-Xent contrastive loss defined above, the objective is formulated as:
\begin{equation}
    \ell_{CL}=\frac{1}{N}\sum_{i=1}^{N}log\frac{\ell(z_{s}^{(i)},z_{a}^{(i)})}{\sum_{k=1}^{N} 1_{[k\ne i]}\ell (z_{s}^{(i)},z_{s}^{(k)})+\ell(z_{s}^{(i)},z_{a}^{(k)})}
\end{equation}
where $z_{s}$ denotes the textual representation obtained by encoding the original SMILES string, and $z_{a}$ denotes the representation obtained by encoding the corresponding LLM-augmented textual description. All other symbols are defined in the same manner as in Eq.\ref{contrastive}.\par
Through this contrastive learning objective, the model is encouraged to associate structural cues derived from SMILES with complementary semantic information provided by the augmented text, enabling the learned representations to simultaneously encode molecular topological information and chemically meaningful semantic attributes.

\begin{figure*}[t]
    \centering
    \includegraphics[width=\textwidth]{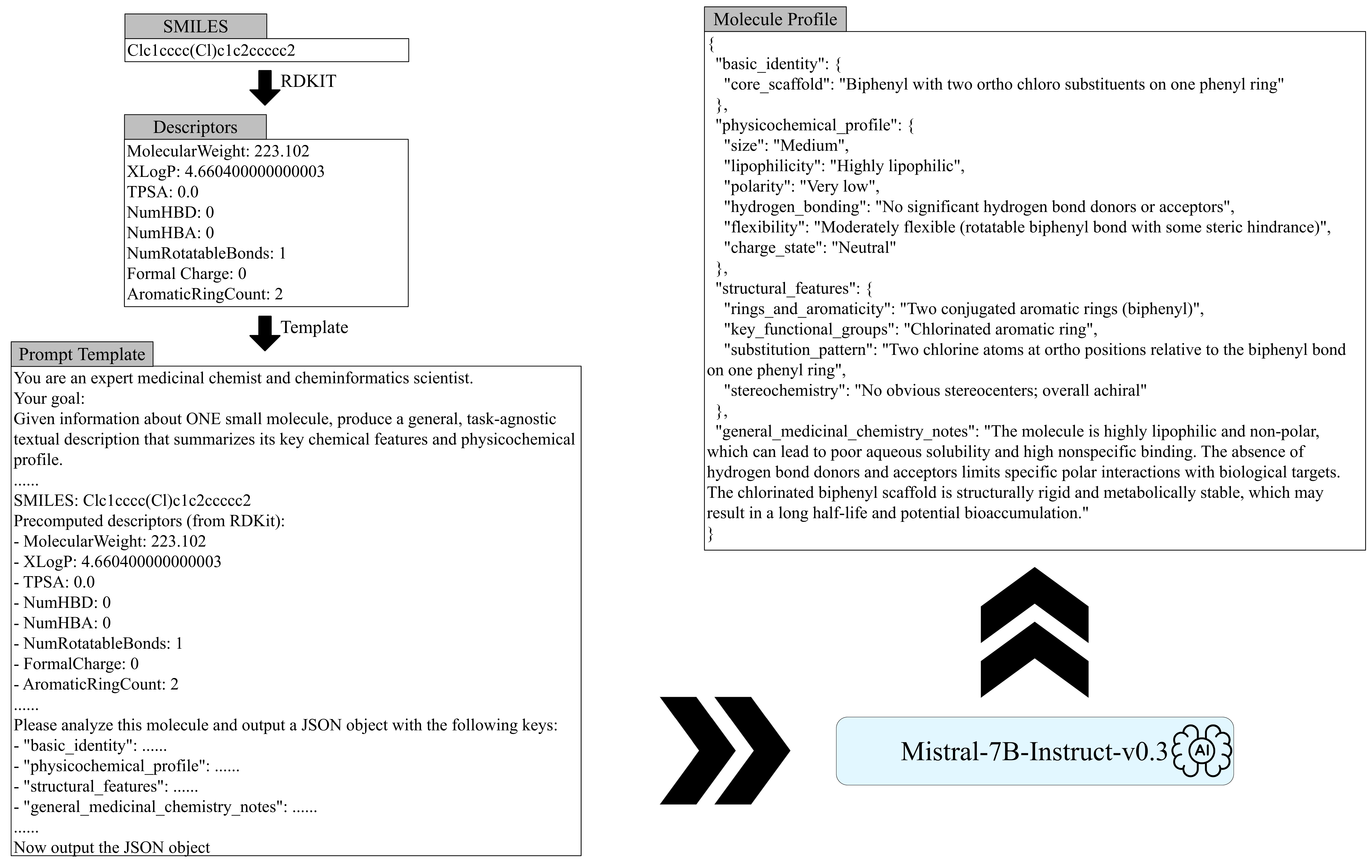}
    \caption{Workflow for enriching SMILES strings with chemically meaningful natural language descriptions generated by Mistral-7B-Instruct-v0.3.}
    \label{fig:Template}
\end{figure*}
\subsection{Multi-Modal Fusion and Fine-Tuning}
After pretraining the modality-specific encoders, we perform multimodal fusion and fine-tuning to obtain a unified molecular representation for downstream property prediction. During this stage, the pretrained graph and text encoders are reused, and their parameters are initialized from the contrastive pretraining stage.\par
For the graph modality, the molecular graph is processed by both the Attentive FP encoder and the Graph Transformer, producing local and global graph representations, respectively. To integrate these complementary structural features, we employ a cross-attention mechanism that aggregates the two representations into a unified graph-level embedding, enabling the model to jointly capture localized substructure motifs and functional group patterns as well as long-range topological dependencies.\par
For the text modality, the original SMILES string and its LLM-enhanced, chemistry-aware textual description are encoded by pretrained DeBERTa model to obtain structural and physicochemical semantic representations. These two textual representations are then fused via cross-attention to form a unified text-level embedding that combines molecular structural information with domain-specific physicochemical semantics.\par
In addition, multiple molecular fingerprints, including MACCS keys \cite{durant2002reoptimization}, PubChem fingerprints \cite{bolton2008pubchem}, and ErG pharmacophore fingerprints \cite{stiefl2006erg}, are introduced as an additional complementary modality. The fingerprint vectors are projected into a latent embedding space through an MLP, yielding a fingerprint-based molecular representation that captures substructure and pharmacophoric patterns.\par
After obtaining modality-specific representations for the graph, text, and fingerprint modalities, we further integrate them using a Dual Cross-attention Mechanism \cite{zhang2024mvmrl}. This module enables bidirectional interactions among heterogeneous modalities, facilitating effective information exchange and alignment across structural, semantic, and fingerprint-derived representations. The output of the Dual Cross-attention module serves as the final molecular representation.\par
Finally, the fused molecular representation is fed into a task-specific MLP to perform molecular property prediction. Through this hierarchical fusion and fine-tuning process, the model effectively combines local and global structural information, physicochemical semantics and fingerprint-based features, resulting in expressive and transferable molecular representations for downstream tasks.

\section{Experiments}
\subsection{Experiments Settings}
\subsubsection{Datasets}

For the self-supervised pretraining stage, we utilize the ZINC15 dataset \cite{sterling2015zinc}, which contains around 330k unlabeled molecules represented by SMILES strings. This large-scale dataset provides diverse molecular structures and is widely used for pretraining molecular representation models. For downstream evaluation, we finetune the pretrained model on ten benchmark datasets from MoleculeNet \cite{wu2018moleculenet}, covering both classification and regression tasks. Specifically, the classification benchmarks include BBBP, BACE, Clintox, Tox21, SIDER, HIV, and ToxCast, while the regression benchmarks consist of FreeSolv, ESOL, and Lipophilicity. These datasets span a wide range of molecular property prediction tasks, including biophysical activity, physiological relevance, and physicochemical property estimation. Detailed statistics and task descriptions of the ten datasets are summarized in Table \ref{tab:datasets}. In addition, we provide LLM-enhanced textual versions for both the ZINC15 pretraining dataset and the ten MoleculeNet finetuning datasets. These textual datasets serve as an additional semantic view during training and are publicly available on our GitHub repository.
\begin{table*}[t]
\centering
\renewcommand{\arraystretch}{1.2}
\caption{Summary of benchmark datasets for molecular property prediction.}
\label{tab:datasets}
\setlength{\tabcolsep}{4pt}
\begin{tabular}{llcccccl}
\toprule
Category & Dataset & Tasks & Molecules & Task Type & Metric & Avg. nodes & Description \\
\midrule
\multirow{2}{*}{Biophysics}
&BACE & 1 & 1513 & Classification & ROC-AUC &34.1 & $\beta$-secretase inhibitory activity\\
&HIV&1 &41127 & Classification & ROC-AUC &25.5 & HIV replication inhibition \\
\midrule
\multirow{5}{*}{Physiology}
& BBBP & 1 & 2053 & Classification & ROC-AUC & 23.9 & Blood-brain barrier permeability\\
& SIDER & 27 & 1427 & Classification & ROC-AUC & 33.6 & Drug-induced side effects\\
& Tox21 & 12 & 7831 & Classification & ROC-AUC & 18.6 & Compound toxicity\\ 
& ClinTox & 2 & 1478 & Classification & ROC-AUC & 26.1 & Clinical toxicity\\
& ToxCast & 617 &8575 & Classification & ROC-AUC & 18.7 & Toxicity screening(HTS)\\
\midrule
\multirow{3}{*} {Physical Chemistry}
& ESOL & 1 & 1128 & Regression & RMSE & 13.3 & Aqueous solubility\\
& FreeSolv & 1 & 642 & Regression & RMSE & 8.7 & Hydration free energy\\
& Lipophilicity & 1 & 4200 & Regression & RMSE &27.0 & Compound lipophilicity(LogD)\\
\bottomrule
\end{tabular}
\end{table*}

\subsubsection{Implementation Details}

For graph-based contrastive learning, we employ a two-layer Attentive FP encoder to capture localized molecular patterns, such as substructure motifs and functional group information. To model global structural dependencies, we adopt a lightweight two-layer Graph Transformer encoder that is designed for efficiently capture long-range topological context across the molecular graph. During contrastive pretraining, the projection heads are applied to the outputs of both graph encoders, and the model is optimized using a contrastive loss with a temperature parameter set to 0.1. The pretraining process is conducted for 100 epochs to ensure stable convergence of the learned representations. For text-based contrastive learning, we adopt a four-layer DeBERTa encoder to model the semantic representations of molecular textual inputs. The official tokenizer provided with DeBERTa is used to tokenize both the original SMILES strings and their corresponding LLM-enhanced textual descriptions, ensuring consistency with the pretrained language model vocabulary. Similar to the graph-based pretraining stage, the text encoder outputs are passed through projection heads and optimized using a contrastive learning objective with the temperature parameter set to 0.1. The text-based contrastive pretraining is also conducted for 100 epochs to ensure sufficient convergence. During the fine-tuning stage, molecular fingerprint features are incorporated through a two-layer MLP to obtain fingerprint-based representations. For all downstream datasets, the model is finetuned for 50 epochs. Consistent with pretraining stages, the Adam optimizer \cite{kingma2014adam} is employed during finetuning, with the initial learning rate set to $1\times10^{-3}$. All experiments were conducted on an NVIDIA RTX 4090 GPU. Regarding data splitting, we follow the random splitting protocol adopted in S-CGIB \cite{lee2025pre}, using a split ratio of $0.6/0.2/0.2$ for training, validation, and testing, and report the average performance over three runs with different random seeds. In addition to random splitting, we also evaluate the model under the scaffold splitting \cite{bemis1996properties}, which is widely regarded as a more challenging and realistic evaluation protocol for molecular property prediction. The detailed results under scaffold splitting are provided in the supplementary information.

\subsubsection{Baselines}
To demonstrate the effectiveness of the proposed method, we compare it with a diverse set of state-of-the-art self-supervised learning baselines for molecular graph representation learning. The experimental results of these baselines are obtained under the same benchmark settings as S-CGIB \cite{lee2025pre}, which we strictly follow in our experiments.These baselines can be broadly categorized into three groups.\par

\begin{itemize}

    \item Predictive self-supervised methods, which learn graph representations via reconstruction or prediction-based pretext tasks without explicit contrastive objectives. The category includes ContextPred \cite{hu2019strategies} and AttrMasking \cite{hu2019strategies}, which predict contextual or masked node attributes, as well as EdgePred \cite{hamilton2017inductive}, GraphFP \cite{subramonian2021motif}, SimSGT \cite{liu2023rethinking}, and MoAMa \cite{inae2023motif} that exploit edge-level fragment-level, or motif-aware masking strategies.
    \item Contrastive learning-based methods, which construct multiple graph views and learn representations by maximizing agreement between positive pairs. Representative approaches in this category include Infomax \cite{velivckovic2018deep}, JOAO \cite{you2021graph}, JOAOv2 \cite{you2021graph} and GraphCL \cite{you2020graph}, which rely on graph augmentations, as well as structure-aware contrastive methods such as GraphLoG \cite{xu2021self}, MICRO-Graph \cite{zhang2021motif} and MGSSL \cite{luong2023fragment}.
    \item Advanced structural modeling methods, which employ more expressive graph architectures for molecular representation learning. This group includes GROVER \cite{rong2020self}, a graph transformer-based pretraining approach, and S-GGIB \cite{lee2025pre}, which incorporates subgraph-level information constraints.
    
\end{itemize}

\subsection{Performance Analysis}
\begin{table*}[t]
\centering
\renewcommand{\arraystretch}{1.2}
\caption{Comparative performance of different methods on classification benchmarks measured by ROC-AUC.}
\label{tab:classification}
\setlength{\tabcolsep}{4pt}
\begin{tabular}{lccccccc}
\toprule
Methods & BACE & BBBP & ClinTox & SIDER & Tox21 & ToxCast & HIV \\
\midrule
ContextPred & $78.39\pm0.58$ & $69.10\pm0.29$ & $55.63\pm1.35$ & $61.83\pm0.60$ & $73.26\pm0.59$ & $63.28\pm0.68$ & $72.04\pm0.48$ \\
AttrMasking & $75.95\pm0.50$ & $67.12\pm0.45$ & $60.11\pm1.19$ & $61.21\pm0.65$ & $73.37\pm0.55$ & $61.66\pm1.20$ & $72.71\pm0.70$ \\
EdgePred    & $74.29\pm1.37$ & $64.73\pm1.10$ & $61.62\pm1.25$ & $60.18\pm0.76$ & $70.32\pm1.62$ & $60.04\pm0.81$ & $70.55\pm1.68$ \\
Infomax     & $77.80\pm0.46$ & $68.39\pm0.64$ & $58.62\pm0.83$ & $59.02\pm0.56$ & $72.66\pm0.16$ & $62.76\pm0.54$ & $73.55\pm0.47$ \\
JOAO        & $74.94\pm1.35$ & $71.63\pm1.11$ & $77.02\pm1.64$ & $63.55\pm0.81$ & $73.67\pm1.06$ & $63.30\pm0.27$ & $77.55\pm1.94$ \\
JOAOv2      & $74.38\pm1.71$ & $71.98\pm0.18$ & $65.22\pm0.75$ & $59.88\pm1.72$ & $73.95\pm1.88$ & $63.12\pm1.90$ & $77.13\pm1.51$ \\
GraphCL     & $77.80\pm0.46$ & $68.39\pm0.64$ & $61.62\pm1.25$ & $61.83\pm0.60$ & $73.26\pm0.59$ & $62.76\pm0.54$ & $73.55\pm0.47$ \\
GraphLoG    & $76.60\pm1.04$ & $66.75\pm0.32$ & $53.76\pm0.95$ & $59.09\pm0.53$ & $71.64\pm0.49$ & $61.53\pm0.35$ & $73.76\pm0.29$ \\
GraphFP     & $80.28\pm3.06$ & $72.05\pm1.17$ & $76.80\pm1.83$ & \boldmath{$65.93\pm3.09$} & $77.35\pm1.40$ & $69.15\pm1.92$ & $75.71\pm1.39$ \\
MICRO-Graph & $63.57\pm1.55$ & $67.21\pm1.85$ & $77.56\pm1.56$ & $60.34\pm0.96$ & $71.79\pm1.70$ & $60.80\pm1.15$ & $76.73\pm1.07$ \\
MGSSL       & $82.03\pm3.79$ & $79.52\pm1.98$ & $75.84\pm1.82$ & $57.46\pm1.45$ & $74.82\pm1.60$ & $63.86\pm1.57$ & $77.45\pm2.94$\\
GROVER       & $81.13\pm0.14$ & $87.15\pm0.06$ & $72.53\pm0.14$ & $57.53\pm0.23$ & $68.59\pm0.24$ & $64.45\pm0.14$ & $75.04\pm0.13$ \\
SimSGT      & $79.75\pm1.28$ & $71.51\pm1.75$ & $74.11\pm1.05$ & $59.74\pm1.32$ & $76.23\pm1.27$ & $65.83\pm0.79$ & $78.13\pm1.07$ \\
MoAMa       & $81.32\pm1.06$ & $85.89\pm0.61$ & $77.11\pm1.67$ & $62.69\pm0.37$ & $78.29\pm0.55$ & $68.01\pm1.07$ & $78.11\pm0.64$ \\
S-CGIB      & $86.46\pm0.81$ & $88.75\pm0.49$ & $78.58\pm2.01$ & $64.03\pm1.04$ & $80.94\pm0.17$ & \boldmath{$70.95\pm0.27$} & \boldmath{$78.33\pm1.34$} \\
Ours        & \boldmath{$87.29\pm0.19$} & \boldmath{$92.44\pm0.48$} & \boldmath{$86.06\pm6.71$} & $60.49\pm0.70$& \boldmath{$81.16\pm0.71$}& $67.49\pm0.56$ & $77.49\pm0.98$\\
\bottomrule
\end{tabular}
\end{table*}

\begin{table*}[t]
\centering
\renewcommand{\arraystretch}{1.2}
\caption{Comparative performance of different methods on regression benchmarks measured by RMSE.}
\label{tab:regression}
\setlength{\tabcolsep}{6pt}
\begin{tabularx}{\textwidth}{l *{3}{>{\centering\arraybackslash}X}}
\toprule
Methods & ESOL & FreeSolv & Lipophilicity \\
\midrule
ContextPred & $2.190\pm0.026$ & $3.195\pm0.058$ & $1.053\pm0.048$ \\
AttrMasking & $2.954\pm0.087$ & $4.023\pm0.039$ & $0.982\pm0.052$ \\
EdgePred    & $2.368\pm0.070$ & $3.192\pm0.023$ & $1.085\pm0.061$ \\
Infomax     & $2.953\pm0.049$ & $3.033\pm0.026$ & $0.970\pm0.023$ \\
JOAO        & $1.978\pm0.029$ & $3.282\pm0.002$ & $1.093\pm0.097$ \\
JOAOv2      & $2.144\pm0.009$ & $3.842\pm0.012$ & $1.116\pm0.024$ \\
GraphCL     & $1.390\pm0.363$ & $3.166\pm0.027$ & $1.014\pm0.018$ \\
GraphLoG    & $1.542\pm0.026$ & $2.335\pm0.052$ & $0.932\pm0.052$ \\
GraphFP     & $2.136\pm0.096$ & $2.528\pm0.016$ & $1.371\pm0.058$ \\
MICRO-Graph & $0.842\pm0.055$ & $1.865\pm0.061$ & $0.851\pm0.073$ \\
MGSSL       & $2.936\pm0.071$ & $2.940\pm0.051$ & $1.106\pm0.077$ \\
GROVER      & $1.237\pm0.403$ & $2.712\pm0.327$ & $0.823\pm0.027$ \\
SimSGT      & $0.932\pm0.026$ & $1.953\pm0.038$ & $0.771\pm0.041$ \\
MoAMa       & $1.125\pm0.029$ & $2.072\pm0.053$ & $1.085\pm0.024$ \\
S-CGIB      & $0.816\pm0.019$ & $1.648\pm0.074$ & $0.762\pm0.042$ \\
Ours        & \boldmath{$0.496\pm0.025$} & \boldmath{$1.153\pm0.053$} & \boldmath{$0.674\pm0.028$}\\
\bottomrule
\end{tabularx}
\end{table*}

\begin{table*}[t]
\centering
\renewcommand{\arraystretch}{1.2}
\caption{Ablation study of single- and bi-modality configurations across six molecular property prediction tasks. The best result on each dataset is bold.}
\label{tab:ablation_study_modal}
\setlength{\tabcolsep}{6pt}

\begin{tabularx}{\textwidth}{l *{6}{>{\centering\arraybackslash}X}}
\toprule
Methods & BACE & BBBP & ClinTox & ESOL & FreeSolv & Lipophilicity \\
\midrule
T       & $0.853$ & $0.911$ & $0.922$ & $0.516$ & $1.460$ & $0.811$ \\
F       & $0.782$ & $0.882$ & $0.857$ & $0.540$ & $1.308$ & $0.699$ \\
G       & $0.782$ & $0.890$ & $0.811$ & $0.544$ & $1.102$ & $0.785$ \\
T+G     & $0.694$ & $0.684$ & $0.862$ & $0.761$ & $1.710$ & $1.083$ \\
T+F     & $0.892$ & $0.917$ & $0.914$ & $0.501$ & $1.111$ & $0.658$ \\
F+G     & $0.841$ & $0.903$ & $0.830$ & $0.621$ & $1.170$ & $0.713$ \\
T+F+G   & \boldmath{$0.875$} & \boldmath{$0.929$} & \boldmath{$0.952$} & \boldmath{$0.478$} & \boldmath{$1.077$} & \boldmath{$0.634$} \\
\bottomrule
\end{tabularx}
\end{table*}

\begin{figure*}[t]
    \centering
    \includegraphics[width=\textwidth]{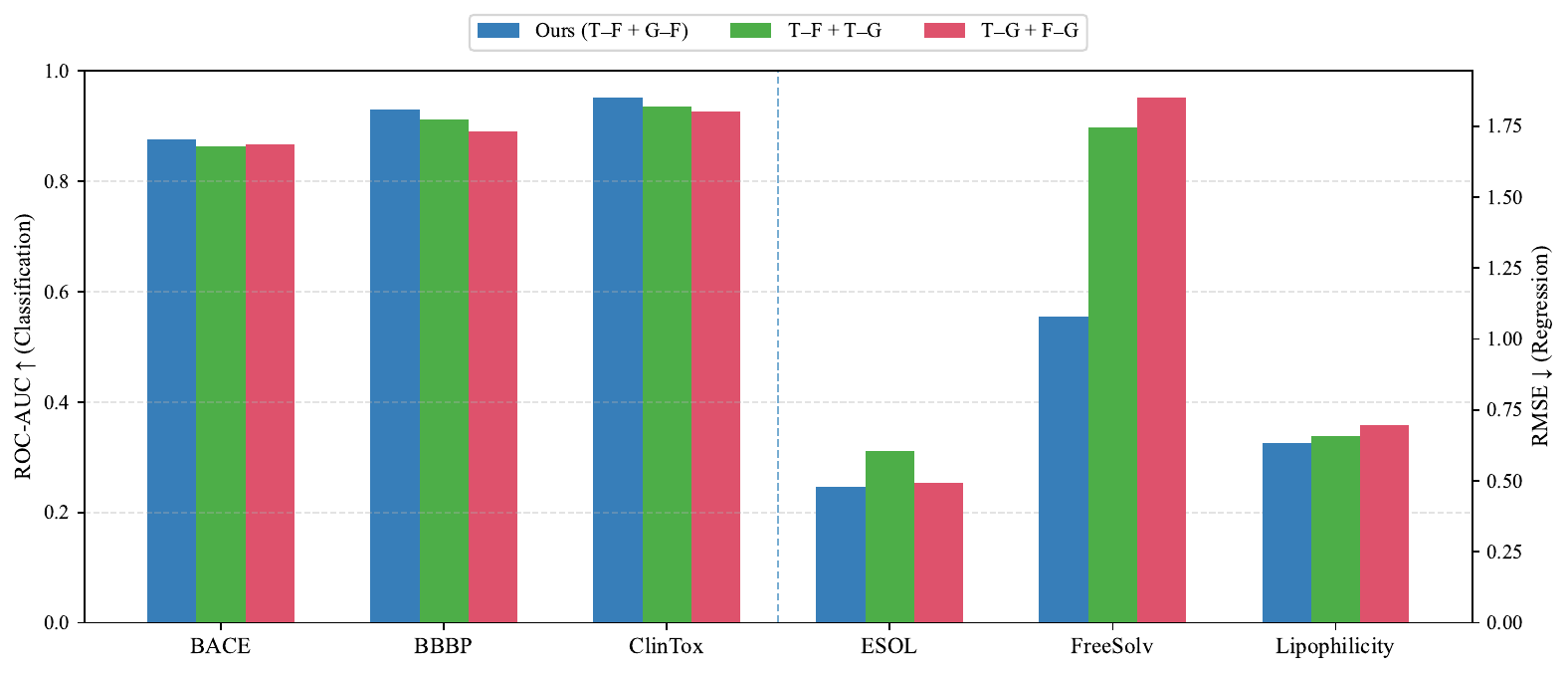}
    \caption{Ablation study of dual cross-attention fusion strategies across six molecular property prediction tasks, reporting ROC-AUC for classification and RMSE for regression.}
    \label{fig:cross-attention}
\end{figure*}

\begin{figure*}
    \centering
    \includegraphics[width=\textwidth]{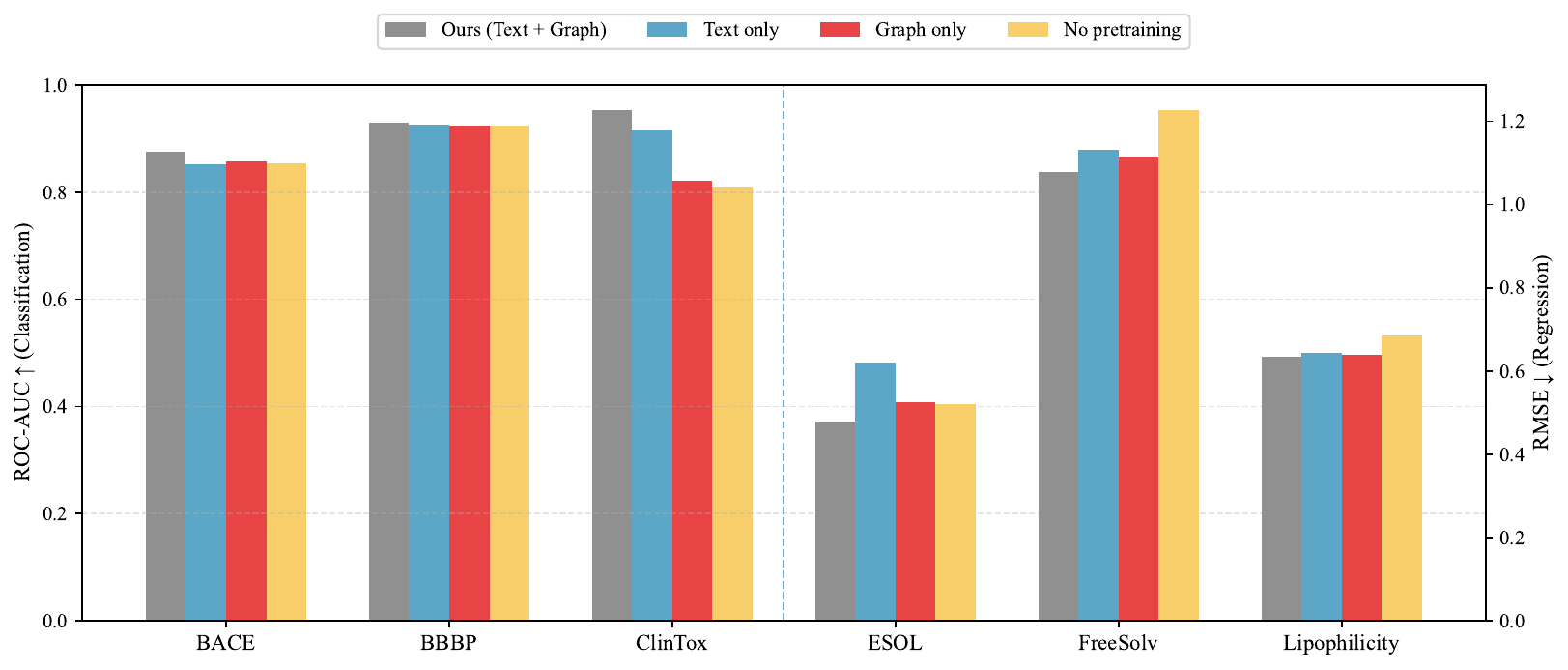}
    \caption{Ablation study of different pretraining initialization strategies across six molecular property prediction tasks, reporting ROC-AUC for classification and RMSE for regression.}
    \label{fig:pretraining}
\end{figure*}

\begin{figure*}
    \centering
    \includegraphics[width=\textwidth]{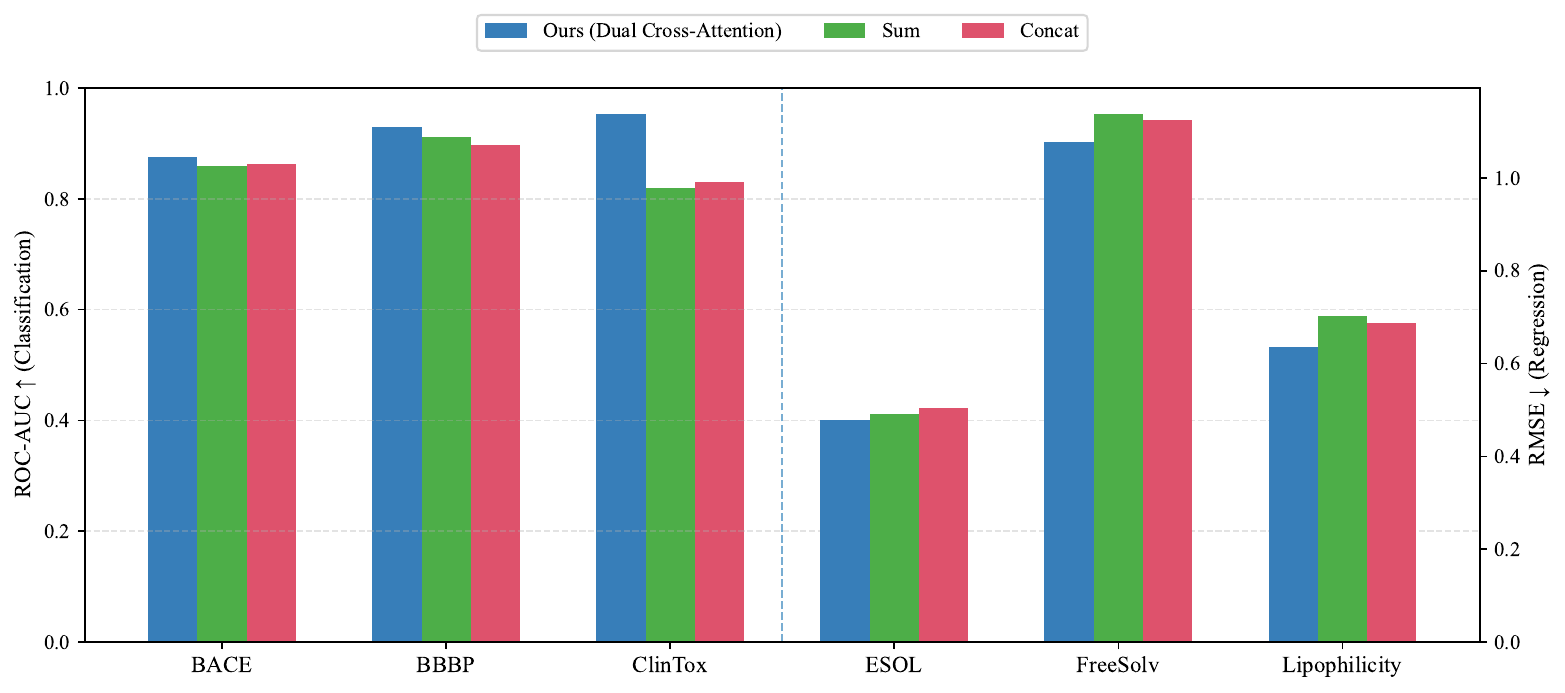}
    \caption{Ablation study of different feature fusion strategies across six molecular property prediction tasks, reporting ROC-AUC for classification and RMSE for regression.}
    \label{fig:strategy}
\end{figure*}

\begin{figure*}[t]
    \centering
    \includegraphics[width=\textwidth]{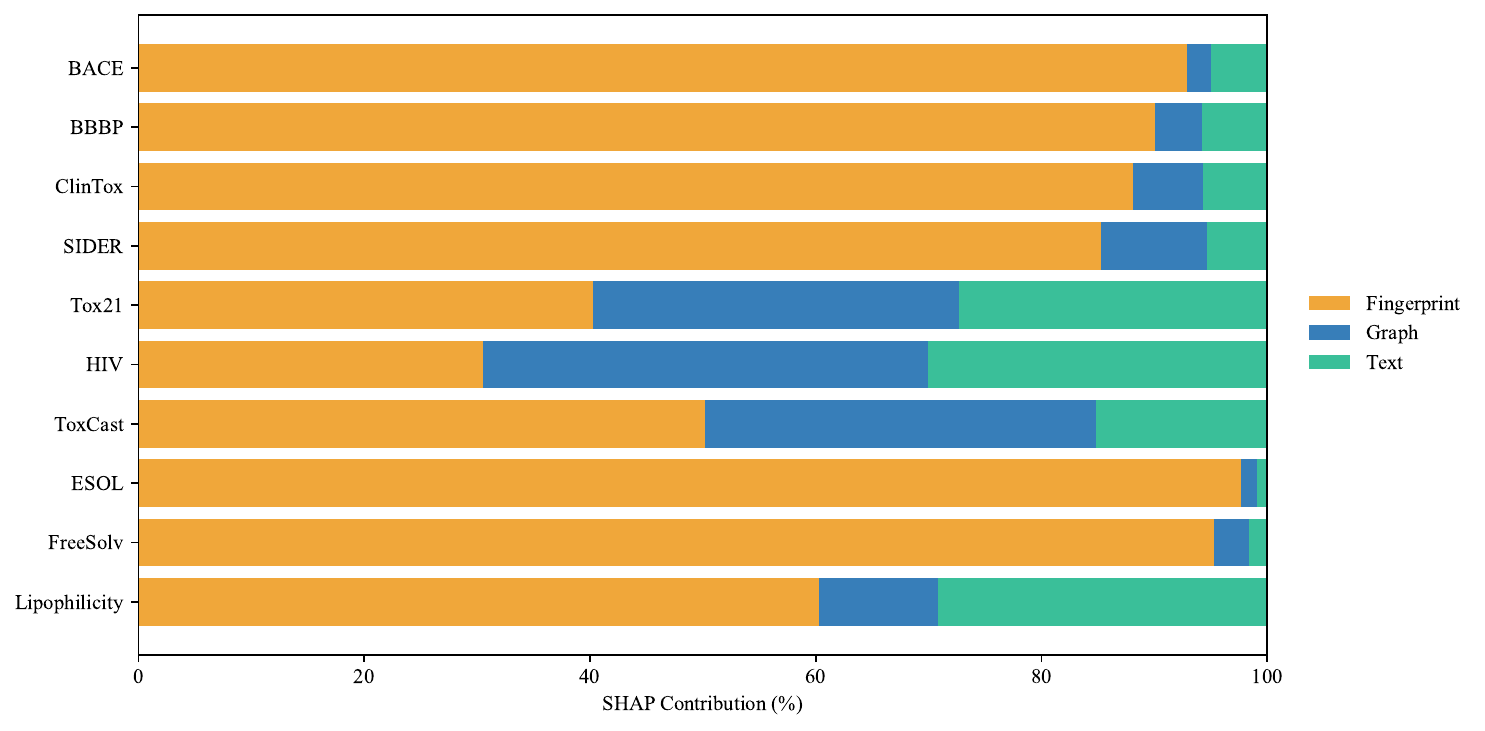}
    \caption{Dataset-dependent modality contribution revealed by SHAP analysis.}
    \label{fig:shap_vis}
\end{figure*}

Table \ref{tab:classification} and Table \ref{tab:regression} report the comparative performance of different baselines on classification and regression benchmarks, respectively. Overall, our proposed method shows robust performance across both classification and regression tasks.\par
On classification tasks, our method consistently improves ROC-AUC scores across challenging benchmarks such as BACE, BBBP, and ClinTox, surpassing strong pretraining approaches including GROVER and S-CGIB. In particular, the performance gains on BBBP and Tox21 demonstrate the effectiveness of our model in capturing complex molecular semantics. For regression tasks, our approach achieves the lowest RMSE on ESOL, FreeSolv, and Lipophilicity, indicating superior regression accuracy compared with existing methods. These results suggest that learned representations are more suitable for fine-grained molecular property prediction. Moreover, the relatively small standard deviations observed on several datasets indicate that our method exhibits stable and robust performance.
\subsection{Ablation Study}
To assess the contribution of different molecular modalities, we perform an ablation study by varying the modality configuration while keeping the training protocol unchanged. We evaluate single-modality, bi-modality, and full-modality settings to examine the individual and joint effects of textual, graph-based, and fingerprint-based representations on downstream tasks.
Table \ref{tab:ablation_study_modal} reports results for three single-modality models and three bi-modality combinations, with the full model shown for reference. Performance is measured using ROC-AUC for classification tasks and RMSE for regression tasks. Overall, bi-modality consistently outperform single-modality counterparts, highlighting the importance of multimodal integration. Graph-based representations are particularly effective for regression tasks, while textual information provides complementary chemical semantics. The full multimodal model achieves the most consistent performance across benchmarks, confirming the benefit of jointly integrating all modalites.\par
In addition to modality-level ablations, we further investigate the impact of different pretraining initialization strategies on downstream performance. In the original model, both text and graph encoders are initialize with pretrained parameters. We then construct three ablation settings by selectively disabling pretrained initialization, including text-only pretraining, graph-only pretraining, and a baseline trained entirely from scratch. The results are illustrated in Figure \ref{fig:pretraining}, which reports performance across the same six benchmark datasets using ROC-AUC for classification tasks and RMSE for regression tasks. This ablation aims to disentangle the individual and complementary contributions of text- and graph-based pretraining, and to assess whether the observed performance gains stem from the proposed pretraining strategy rather than architectural design alone. As shown in Figure \ref{fig:pretraining}, full pretraining of both text and graph encoders achieves the best overall performance. Using only text or graph pretraining results in moderate degradation, while training from scratch leads to a more pronounced drop, especially on regression tasks. These results highlight the complementary contributions of text and graph pretraining. To further validate the effectiveness of contrastive learning, we additionally report single-modality ablations for text-only and graph-only models with and without pretrained initialization in the Supplementary information.\par
Complementary to the pretraining ablation, we investigate the effect of different aggregation strategies for multi-modal feature fusion. Our original model employs a dual cross-attention mechanism for multi-modal feature aggregation. For ablation purposes, we substitute this module with standard aggregation operations, including summation and concatenation, while keeping all other components unchanged. As shown in Figure \ref{fig:strategy}, the dual cross-attention-based aggregation consistently outperforms simple summation and concatenation across both classification and regression benchmarks. This result indicates that explicitly modeling cross-modal interactions is more effective than parameter-free aggregation strategies for multi-modal feature fusion.\par 
Finally, we conduct an ablation study on different modality pairing strategies in the dual cross-attention framework to assess their impact on multi-modal feature fusion. As shown in Figure \ref{fig:cross-attention}, the original fingerprint-centered pairing strategy yields the most consistent performance across benchmarks, while alternative pairing schemes result in performance degradation, highlighting the importance of appropriate modality interaction design.

\subsection{Interpretability Analysis}

\begin{figure*}[t]
    \centering
    \includegraphics[width=\textwidth]{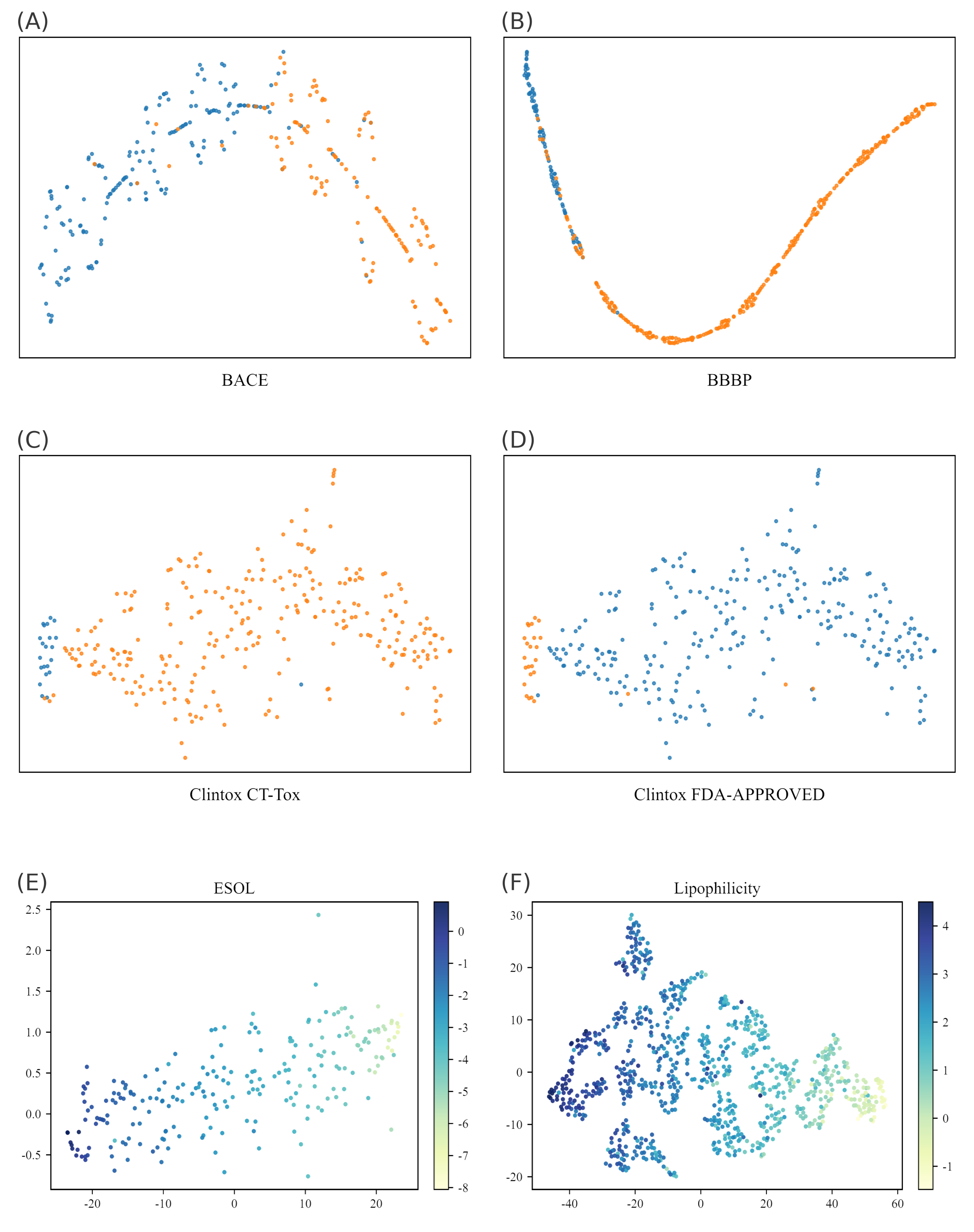}
    \caption{t-SNE visualizations of learned molecular representations on the test sets.(A) BACE. (B) BBBP. (C) ClinTox(CT-Tox). (D) Clintox(FDA-APPROVED). (E) ESOL. (F). Lipophilicity. For classification tasks, different colors indicate different class labels, while for regression tasks, the color gradient reflects the corressponding target property values. The visualizations illustrate the organization of molecular embeddings learned by the model in a two-dimensional space.}
    \label{fig:Tsne}
\end{figure*}
To interpret the decision-making process of the proposed multimodal framework and quantify the contribution of each modality, we perform an interpretability analysis using SHAP (SHapley Additive exPlanations) \cite{lundberg2017unified}. SHAP Provides model-agnostic attributions based on Shapley values, enabling fair and additive estimation of feature contributions. In this work, feature-level SHAP values are aggregated at the modality level to assess the relative importance of molecular fingerprints, molecular graphs, and texts in the final prediction.\par
The modality-level SHAP contributions across multiple benchmark datasets are shown in Figure \ref{fig:shap_vis}. The result exhibit clear task-dependent differences. For datasets such as BACE, BBBP, Clintox, SIDER, ESOL, and FreeSolv, fingerprint representations dominate the predictions, contributing the majority of the overall importance, while graph- and text-based modalities play a minor role. This suggests that, for these tasks, compact substructure patterns encoded in fingerprints are sufficient to capture the key determinants of molecular properties. In contrast, more complex datasets such as Tox21 and HIV display a more balanced contribution across modalities, including that accurate prediction requires the joint use of structural, topological, and semantic information. The Toxcast dataset shows a similar trend, with fingerprints and graph representations jointly contributing most of the predictive signal, and textual information providing complementary support. For Lipophilicity, the text modality exhibits a relatively higher contribution, highlighting the importance of physicochemical information beyond purely structural descriptors for this task.\par
Overall, this analysis demonstrates that the proposed framework adaptively leverages different modalities depending on the prediction task, rather than relying on a single information source. The SHAP-based results provide quantitative evidence for the effectiveness of multi-modal integration, particularly for challenging molecular property prediction problems where heterogeneous information sources are essential.\par
To gain an intuitive understanding of the learned molecular representations, we employ t-SNE \cite{maaten2008visualizing}, a widely used nonlinear dimensionality reduction technique for visualizing high-dimensional data in a low-dimensional space. t-SNE maps samples into a two-dimensional embedding by preserving local neighborhood similarities, enabling the inspection of structural patterns and class separability in the learned feature space.\par
Figure \ref{fig:Tsne} presents the t-SNE visualizations of the model representations on the test sets of several benchmark datasets, obtained under the random splitting protocol with a ratio of 0.6/0.2/0.2. For classification tasks (BACE, BBBP, and ClinTox), different colors denote different class labels, while for regression tasks (ESOL and Lipophilicity), the color intensity reflects the magnitude of the target property In addition, we also provide t-SNE visualization of models trained under the scaffold splitting protocol in the supplementary material.\par 
Overall, the visualizations indicate that the learned representations exhibit meaningful organization in the embedding space. In several datasets, samples with similar labels or property values tend to form coherent clusters or follow smooth structural trends, suggesting that the model captures task-relevant molecular characteristics. Those observations provide qualitative evidence that the proposed representation learning framework produces informative and discriminative embeddings.
\section{Conclusion}
In this work, we proposed LGM-CL, a local-global multimodal contrastive learning framework for molecular property prediction. By jointly modeling local functional group and global molecular topology from molecular graphs, and aligning them with chemistry-aware textual representations derived from SMILES and augmented texts, the proposed method learns transferable and chemically meaningful molecular embeddings. During fine-tuning, molecular fingerprints are further integrated through dual cross-attention module to enhance downstream prediction. Extensive experiments on MoleculeNet benchmarks demonstrate that LGM-CL achieves consistent and competitive performance across both classification and regression tasks. Ablation and interpretability analyses further validate the complementary contributions of different modalities. These results highlight the effectiveness of unified local-global and multimodal representation learning for molecular property prediction.
\section{Key points}
\begin{itemize}
    \item We propose a local-global multimodal contrastive learning framework that jointly models fine-grained local chemical patterns and global molecular structural dependencies.
    \item A chemistry-aware text augmentation strategy is introduced to align SMILES with enriched physicochemical semantics through contrastive learning in a self-supervised manner.
    \item A dual cross-attention fusion mechanism effectively integrates graph, text and fingerprint representations for unified molecular representation learning.
    \item Extensive experiments on MoleculeNet benchmarks demonstrate consistent and robust performance gains across both classification and regression tasks, supported by comprehensive ablation and interpretability analysis.
\end{itemize}

\section{Competing interests}
The authors declare no competing financial interest.
\section{Data availability}
The datasets used in the experiments, along with their corresponding text-augmented versions, are publicly available at https://github.com/lhb0189/LGM-CL.
\section{Acknowledgments}
This work is supported by National Natural Science Foundation of China (Grant: 11101071). Supported by Center for HPC, University of Electronic Science and technology.
\bibliographystyle{plain}



\end{document}